\documentclass[a4paper,conference]{IEEEtran}

\usepackage{times}
\usepackage{epsfig}
\usepackage{graphicx}
\usepackage{amsmath}
\usepackage{amssymb}
\usepackage{blindtext}
\usepackage{framed}
\usepackage[breaklinks=true,bookmarks=false]{hyperref}
\usepackage{mathtools}
\usepackage{dirtytalk}
\usepackage{subfigure}
\usepackage{url}

\newcommand{\R}{\mathbb{R}}

\DeclarePairedDelimiter\norm{\lVert}{\rVert}%

\let\oldnorm\norm
\def\norm{\@ifstar{\oldnorm}{\oldnorm*}}
\makeatother

\begin{document}


\title{Spatio-Temporal Event Segmentation and Localization for Wildlife Extended Videos}


\author{\IEEEauthorblockN{Ramy Mounir\IEEEauthorrefmark{1},
Roman Gula\IEEEauthorrefmark{2},
J\"orn Theuerkauf\IEEEauthorrefmark{2} and
Sudeep Sarkar\IEEEauthorrefmark{1}}
\IEEEauthorblockA{\IEEEauthorrefmark{1}
University of South Florida, 
Tampa, FL, USA
\{ramy, sarkar\}@usf.edu}
\IEEEauthorblockA{\IEEEauthorrefmark{2}
Museum and Institute of Zoology,
Polish Academy of Sciences,
Warsaw, Poland
\{rgula, jtheuer\}@miiz.eu }

}

\maketitle

\begin{abstract}
Using offline training schemes, researchers have tackled the event segmentation problem by providing full or weak-supervision through manually annotated labels or self-supervised epoch-based training. Most works consider videos that are at most 10's of minutes long. We present a self-supervised perceptual prediction framework capable of temporal event segmentation by building stable representations of objects over time and demonstrate it on long videos, spanning several days. The approach is deceptively simple but quite effective. We rely on predictions of high-level features computed by a standard deep learning backbone. For prediction, we use an LSTM, augmented with an attention mechanism, trained in a self-supervised manner using the prediction error. The self-learned attention maps effectively localize and track the event-related objects in each frame.  The proposed approach does not require labels. It requires only a single pass through the video, with no separate training set.  Given the lack of datasets of very long videos, we demonstrate our method on video from 10 days (254 hours) of continuous wildlife monitoring data that we had collected with required permissions. We find that the approach is robust to various environmental conditions such as day/night conditions, rain, sharp shadows, and windy conditions. For the task of temporally locating events, we had an 80\% recall rate at 20\% false-positive rate for frame-level segmentation. At the activity level, we had an 80\% activity recall rate for one false activity detection every 50 minutes. We will make the dataset, which is the first of its kind, and the code available to the research community.
\end{abstract}


\section{Introduction}
One of the tasks involved in wild-life monitoring, or even in the video monitoring of other contexts, is detecting significant events in very long videos, spanning several days. The goal is to flag temporal segments and highlight possible events in the video snippets flagged, i.e., spatial and temporal localization of possible events, such as bird leaving the nest, bird walking into a nest, or the bird building a nest. One cannot rely on very low-level features for this task as they may change due to environmental conditions. We need high-level features that are sufficient to capture object-level representations and a model to capture these features' temporal evolution over time. There are very few works in the literature that show performance on video spanning several days.

\begin{figure}
\centering
\includegraphics[width = \linewidth]{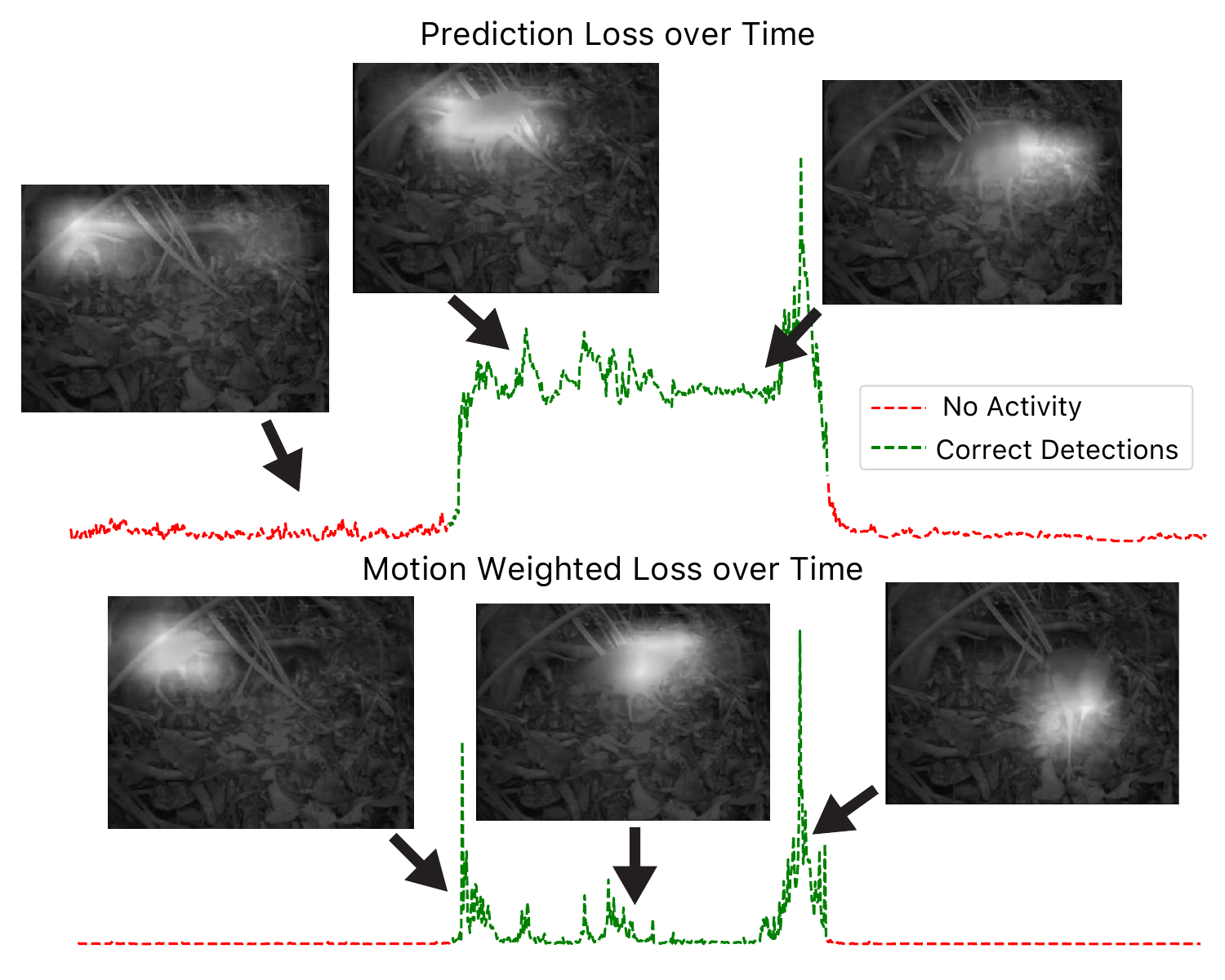}
\caption{Plots of the two kinds of errors before, during, and after an activity: (top) feature prediction loss over the frames, (bottom) motion weighted feature prediction loss over the frames. Errors for some selected frames are shown for both plots, overlaid with the corresponding attention map.} \label{fig:event}
\end{figure}

Event segmentation research has largely focused on offline epoch-based training methods which requires training the model on the entire training dataset prior to testing its performance. This poses a challenge for many real world applications, where the entire dataset is simply non-existent and has to be collected sequentially in a stream over time \cite{online}. Our training scheme completely disregards datapoints after being processed by the network. Training and inference are done simultaneously, alleviating the need for epoch-based training in order to appeal to more practical applications and reduce training time.

Our framework follows key ideas from the perceptual prediction line of work in cognitive psychology \cite{zacks2001perceiving, zacks2001, zacks2007}. Research has shown that \say{event segmentation is an ongoing process in human perception, which helps form the basis of memory and learning}. Humans can identify event boundaries, in a purely bottom up fashion, using a biological perceptual predictive model which predicts future perceptual states based on the current perceived sensory information. Experiments have shown that human perceptual system identifies event boundaries based on the appearance and motion cues in the video \cite{zacks_movement, speer_movement}. Our model implements this perceptual predictive framework and introduces a motion weighted loss function to allow for the localization and processing of motion cues.

Our approach uses a feature encoding network to transform low level perceptual information to higher level feature representation. The model is trained to predict the future perceptual encoded input and signal an event if the prediction is significantly different from the actual features of the next perceived input. The prediction signal also incorporates a higher level representation of the movement cues within frames.

\textbf{Novel contributions: } To the best of our knowledge, we are among the first to (1) introduce the attention-based mechanism to temporal event segmentation models, allowing the model to localize the event in each processed frame, in a purely self-supervised manner, without the need for labels or training data; (2) introduce the idea of motion weighted loss function to stabilize the attention maps that works even when the object of interest does not move; and (3) evaluate and report the performance of temporal segmentation on a remarkably long dataset (over ten days of continuous wildlife monitoring).


\section{Relevant Work}

\textbf{Supervised temporal event segmentation} uses direct labelling (of frames) to segment videos into smaller constituent events. Fully supervised models are heavily dependent on vast amount of training data to achieve good segmentation results. Different model variations and approaches have been tested, such as using an encoder-decoder temporal convolutional network (ED-TCN) \cite{TCN}, or a spatiotemporal CNN model \cite{spatiotemporal_CNN}. To alleviate the need for expensive direct labelling, weakly supervised approaches \cite{RNN, weakly_ordered, weakly_soft, weakly_connectionist} have emerged with an attempt to use metadata (such as captions or narrations) to guide the training process without the need for explicit training labels \cite{weakly_speech, weakly_narrated}. However, such metadata are not always available as part of the dataset, which makes weakly supervised approaches inapplicable to most practical applications.

\textbf{Self-supervised temporal event segmentation} attempts to completely eliminate the need for annotations \cite{unsupervised_complex, unsupervised_predicting}. Many approaches rely heavily on higher level features clustering of frames to sub-activities \cite{unsupervised_clustering, unsupervised_clustering_2}. The performance of the clustering algorithms in unsupervised event segmentation is proportional to the performance of the embedding/encoding model that transforms frames to higher level feature representations. Clustering algorithms can be highly computationally expensive depending on the number of frames to be clustered. Recent work \cite{perceptual_event_segmentation} uses a self-supervised perceptual predictive model to detect event boundaries; we improve upon this model to include attention unit, which helps the model focus on event-causing objects. Other work \cite{event_boundaries} uses a self-supervised perceptual prediction model that is refined over significant amount of reinforcement learning iterations.

\textbf{Frame predictive models} have attempted to provide accurate predictions of the next frame in a sequence \cite{PredNet,PredRnn++, HPNet, Long_Term_Prediction, Physicial_interaction_prediction}; however, these models are focusing on predicting future frames in raw pixel format. Such models may generate a prediction loss that only captures frame motion difference with limited understanding of higher level features that constitutes event boundaries.

\textbf{Attention units} have been applied to image captioning \cite{show_attend_tell}, and natural language processing \cite{attention_is_all_you_need, bahdanau_attention, luong_attention, bert, xlnet} fully supervised applications. Attention is used to expose different temporal - or spatial - segments of the input to the decoding LSTM at every time step using fully supervised model architectures. We use attention in a slightly different form, where the LSTM is decoded only once (per input frame) to predict future features and uses attention weighted input to do so. Unlike \cite{show_attend_tell, attention_is_all_you_need, bahdanau_attention, luong_attention, bert, xlnet}, our attention weights and biases are trained using an unsupervised loss functions. 

Recent work \cite{action_local} has used the prediction loss, with the assistance of region proposal networks (RPNs) and multi-layer LSTM units, to localize actions. We eliminate the need for RPNs and multi-layer LSTM units by extracting Bahdanau \cite{bahdanau_attention} attention weights prior to the LSTM prediction layer, which allows our model to localize objects of interest, even when stationary. From our experiments, we found out that prediction loss attention tends to fade away as moving objects become stationary, which makes its attention map more similar to results extracted from background subtraction or optical flow. In contrast, our model proves to be successful in attending to moving and stationary objects despite variations in environmental conditions, such as moving shadows and lighting changes, as presented in the supplementary video \footnote{Available at \url{https://ramyamounir.github.io/projects/EventSegmentation/}}.


\section{Methodology} \label{sec:method}

\begin{figure}[h]
\centering
\includegraphics[width = 0.70\columnwidth]{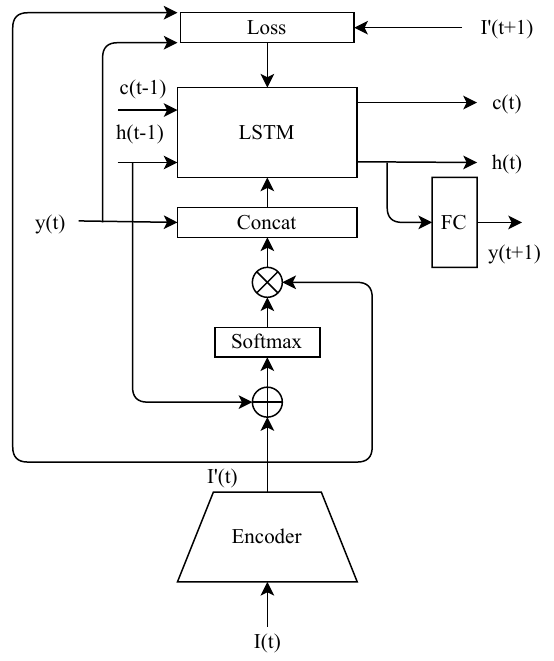}
\caption{The architecture of the self-learning, perceptual prediction algorithm. Input frames from each time instant are encoded into high-level features using a deep-learning stack, followed by an attention overlay that is based on inputs from previous time instant, which is input to an LSTM. The training loss is composed based on the predicted and computed features from current and next frames.} \label{fig:Arch}
\end{figure}

The proposed framework is inspired by the works of Zacks {\it et al.} on perceptual prediction for events segmentation \cite{zacks2007}. The proposed architecture, summarised in Figure \ref{fig:Arch}, can be divided into several individual components. In this section, we explain the role of each component starting by the encoder network and attention unit in sections \ref{sec:encoding} \& \ref{sec:attn}, followed by a discussion on the recurrent predictive layer in section \ref{sec:FPL}. We also introduce the different loss functions (section \ref{sec:loss}) used for self-supervised learning as well as the adaptive thresholding function (section \ref{sec:EG}). We conclude by providing the implementation details (section \ref{sec:details}) used to generate the segmentation results in section \ref{sec:results}.

\subsection{Input Encoding} \label{sec:encoding}

The raw input images are transformed from pixel space into a higher level feature space by utilizing an encoder (CNN) model. This encoded feature representation allows the network to extract features of higher importance to the task being learned. We denote the output of the CNN layers by $I'_t = f(I_t, \theta_e)$ where $\theta_e$ is the learnable weights and biases parameters and $I_t$ is the input image. The encoder network transforms an input image with dimensions $W \times H \times D$ to output features with dimensions $N \times N \times M$, where $N \times N$ is the spatial dimensions and $M$ is the feature vector length.

\subsection{Attention Unit} \label{sec:attn}
In this framework, we utilize Bahdanau attention \cite{bahdanau_attention} to spatially localize the event in each processed frame. The attention unit receives as an input the encoded features and outputs a set of attention weights ($A_t$) with dimensions $N \times N \times 1$. The hidden feature vectors ($h_{t-1}$) from the prediction layer of the previous time step is used to calculate the output set of weights using Equation \ref{eqn:attn}, expressed visually in Figure \ref{fig:Arch}.

\begin{equation} \label{eqn:attn}
    A_t = \gamma(\, \Gamma (\, \varphi( \Gamma(h_{t-1}) + \Gamma(I'_t) )\, )\,)
\end{equation}

Where $\varphi$ represents hyperbolic tangent ($\tanh$) function, $\Gamma$ represents a single fully connected neural network layer and $\gamma$ represents a softmax function. The weights ($A_t$) are then multiplied by the encoded input feature vectors ($I'_t$) to generate the masked feature vectors ($I''_t$). Attention mask is extracted from $A_t$, linearly scaled and resized, then overlayed on the raw input image ($I_t$) to produce the attention visuals shown in Figures \ref{fig:samples1} \& \ref{fig:samples2}.

\subsection{Future Prediction Layer} \label{sec:FPL}
The process of future prediction requires a layer capable of storing a flexible internal state of the previous frames. For this purpose, we use a recurrent layer, specifically long-short term memory cell (LSTM) \cite{lstm}, which is designed to output a future prediction based on the current input and a feature representation of the internal state. More formally, the LSTM cell can be described using the function $h_t = g(I'_t, W_{lstm}, h_{t-1})$, where $h_t$ and $h_{t-1}$ are the output hidden state and previous hidden state respectively, $I'_t$ the encoded input features at time step $t$ and $W_{lstm}$ is a set of weights and biases vectors  controlling the internal state of the LSTM. The input to the LSTM can be formulated as:





\begin{equation} \label{eqn:lstm_inp}
    \Gamma(\Gamma(h_{t-1}) \oplus I''_t)
\end{equation}

where $\Gamma$ is a single fully connected layer, $I''_t$ is the masked encoded input feature vector and $h_{t-1}$ is the hidden state from the previous time step. The symbol $\oplus$ represents vectors concatenation.

\subsection{Loss Function} \label{sec:loss}
The perceptual prediction model aims to train a model capable of predicting the feature vectors of the next time step. We define two different loss functions, prediction loss and motion weighted loss.

\paragraph{Prediction Loss}
This function is defined as the L2 Euclidean distance loss between the output prediction and the next frame encoded feature vectors $I'_t$.

\begin{equation} \label{eqn:PL}
    e_t = ||(I'_{t+1}-y'_t )||^2\\
\end{equation}

\paragraph{Motion Weighted Loss}
This function aims to extract the motion related feature vectors from two consecutive frames to generate a motion dependent mask, which is applied to the prediction loss. The motion weighted loss function allows the network to benefit from motion information in higher level feature space rather than pixel space. This function is formally defined as: 

\begin{equation} \label{eqn:MW}
\begin{split}
    e_t =  ||(I'_{t+1} - y'_t )^{\odot 2} \odot (I'_{t+1} - I'_t)^{\odot 2}||^2\\
\end{split}
\end{equation}

where $\odot$ denotes Hadamard (element-wise) operation.


\begin{figure*}[h]
\centering
\includegraphics[width = 0.87 \linewidth]{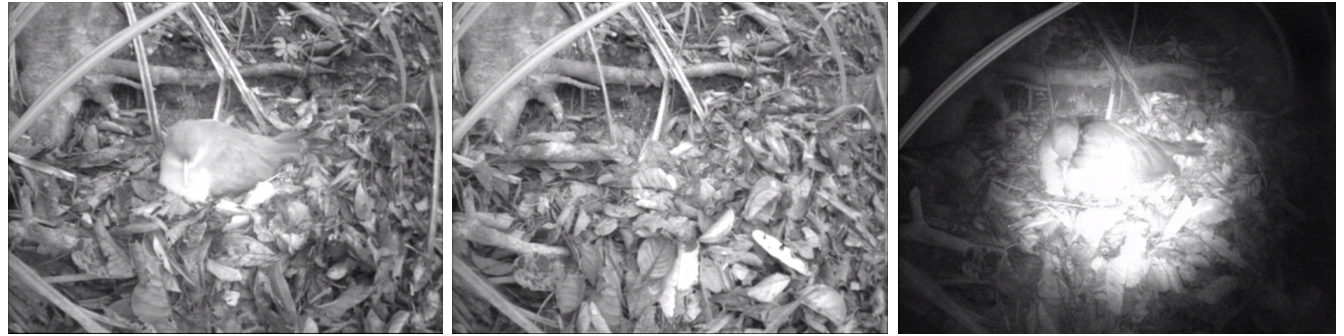}
\caption{Samples of images from the Kagu bird wildlife monitoring dataset} \label{fig:samples}
\end{figure*}

\subsection{Error Gate} \label{sec:EG}
The error gating function receives, as an input, the error signal defined in section \ref{sec:loss}, and applies a thresholding function to classify each frame. In this framework, we define two types of error gating functions. A simple threshold function $f(e_t, \psi)$ and an adaptive threshold function $f(\{e_{t-m} \dots e_t\}, \psi)$. Equation \ref{eqn:conv} formally defines the smoothing function for the adaptive error gating implementation. Both error gating functions use Equation \ref{eqn:thresh} to threshold the error signal. Equations \ref{eqn:conv} \& \ref{eqn:thresh} apply the smoothing function to the full loss signal for analyses purposes; however, the convolution operation can be reduced to element-wise multiplication to calculate a single smoothed value at time step $t$.

\begin{equation} \label{eqn:conv}
\begin{split}
\begin{rcases}
    &e = \{e_{t-m} \dots e_t\} \in \R^m\\
    &e_s = e - [e \, \circledast \, [\{\frac{1}{n} \dots \frac{1}{n}\} \in \R^n]\,] \\
\end{rcases}
n < m
\end{split}
\end{equation}

\begin{equation} \label{eqn:thresh}
        f(e_s(t))= 
\begin{cases}
    1,& \text{if } e_s(t) \geq \psi \\
    0,& \text{otherwise}
\end{cases}
\end{equation}

where $\circledast$ represents a 1D convolution operation.

\subsection{Implementation Details} \label{sec:details}

In our experiments, we use an Inception V3 \cite{inception} encoding model (trained on the ImageNet dataset) to transform input images from raw pixel representation to a higher level feature representation. We freeze the model's parameters (weights and biases) and remove the last layer. The output is a $8 \times 8 \times 2048$ feature tensor, which we reshape to $64 \times 2048$ feature vectors. Each 2048 feature vector requires one LSTM cell for future feature prediction. In other words, the encoded input frame ($I'_t$) is provided with 64 LSTM cells, each processing a 2048 features vector (hidden state size) simultaneously. We use a 0.4 drop rate (recurrent dropout) on the hidden states to prevent overfitting, which may easily occur due to the stateful LSTM nature of the model and the dataset size. LSTMs' Hidden states are initialized to zero. Teacher forcing \cite{teacher} approach is utilized by concatenating the weighted encoded input image ($I''_t$) with the encoded input image ($I'_t$), instead of concatenating it with its prediction from the previous time step ($y_{t-1}$). A single optimization step is performed per frame, Adam optimizer is used with a learning rate of $1e^{-8}$ for the gradient descent algorithm. The dataset is divided into four equal portions and trained on four Nvidia GTX 1080 GPUs simultaneously.


\section{Experimental Evaluation} \label{sec:results}
In this section, we present the results of our experiments for our approach defined in section \ref{sec:method}. We begin by defining the wildlife extended video dataset used for testing, followed by explaining the evaluation metrics used to quantify performance. We discuss the model variations evaluated and conclude by presenting quantitative and qualitative results in sections \ref{sec:quan} \& \ref{sec:qual}.

\begin{figure*}[h]
\centering
\includegraphics[width = 0.86 \linewidth]{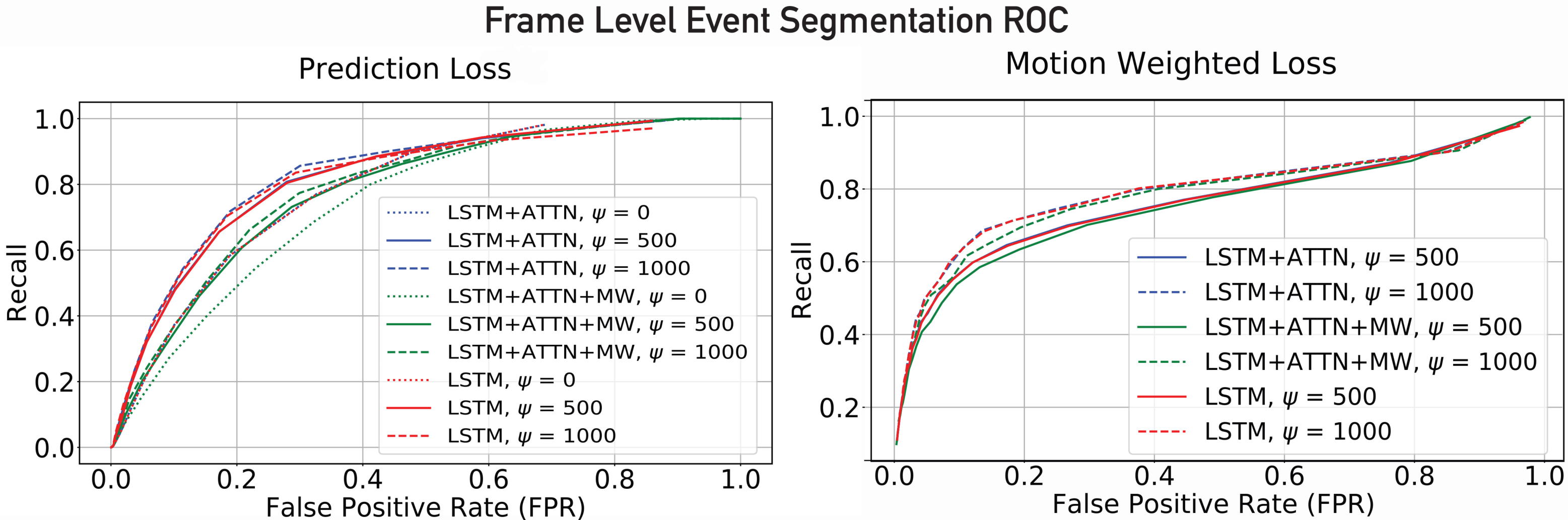}
\caption{Frame-level event segmentation ROCs when activities are detected based on simple thresholding of the prediction and motion weighted loss signals. Plots are shown for different ablation studies.} \label{fig:roc_1}
\end{figure*}

\begin{figure*}[h]
\centering
\includegraphics[width = 0.86 \linewidth]{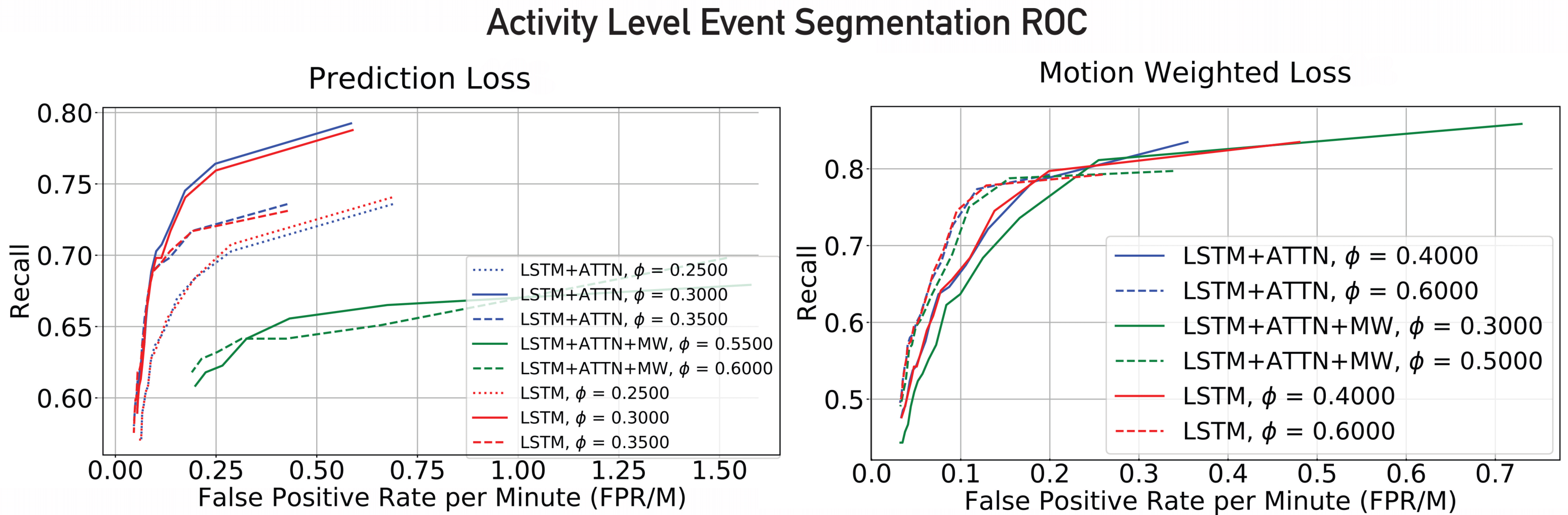}
\caption{Activity-level event segmentation ROCs when activities are detected based on simple thresholding of the prediction and motion weighted loss signals. Plots are shown for different ablation studies.} \label{fig:roc_2}
\end{figure*}

\subsection{Dataset}
We analyze the performance of our model on a wildlife monitoring dataset. The dataset consists of 10 days (254 hours) continuous monitoring of a nest of the Kagu, a flightless bird of New Caledonia. The labels include four unique bird activities, \{feeding the chick, incubation/brooding, nest building while sitting on the nest, nest building around the nest\}. Start and end times for each instance of these activities are provided with the annotations. We modified the annotations to include walk in and walk out events representing the transitioning events from an empty nest to incubation and vice versa. Our approach can flag the nest building (on and around the nest), feeding the chick, walk in and out events. Other events based on climate, time of day, lighting conditions are ignored by our segmentation network. Figure \ref{fig:samples} shows a sample of images from the dataset.

\begin{figure*}[h]
\centering
\includegraphics[width = 0.86 \linewidth]{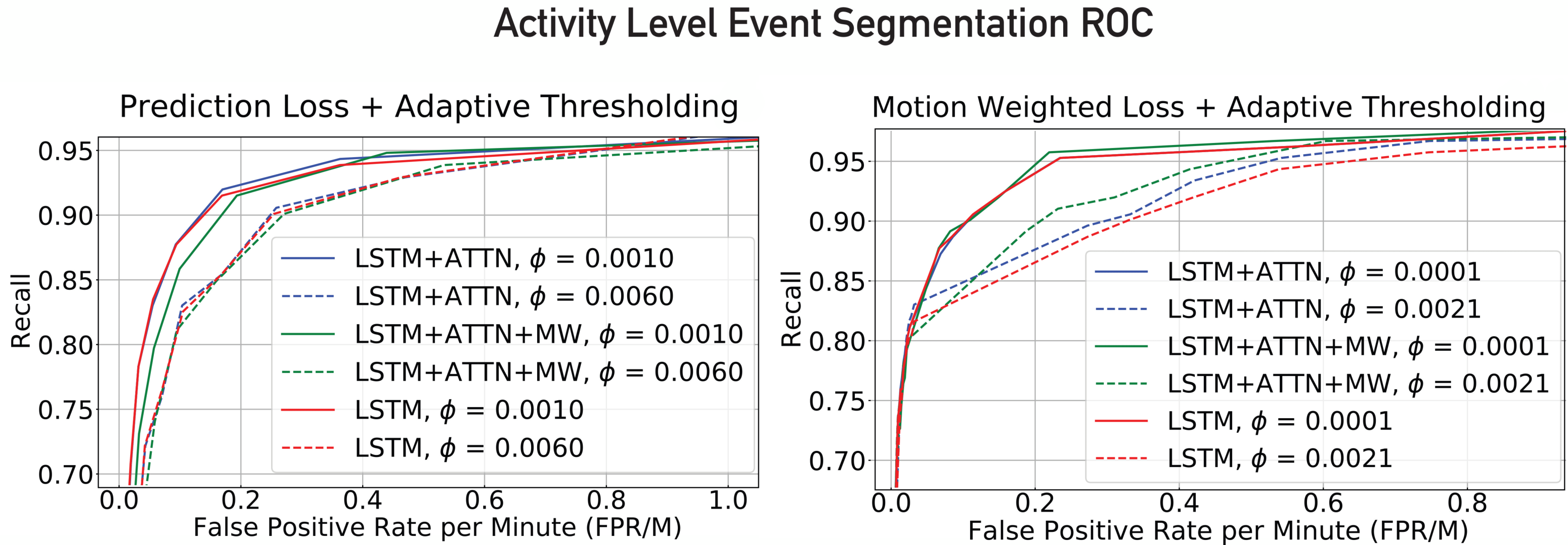}
\caption{Activity-level event segmentation ROCs when activities are detected based on adaptive thresholding of the prediction and motion weighted loss signals. Plots are shown for different ablation studies. } \label{fig:roc_3}
\end{figure*}

\subsection{Evaluation Metrics}
We provide quantitative ROC results for both frame level (figure \ref{fig:roc_1}) and activity level (Figures \ref{fig:roc_2} \& \ref{fig:roc_3}) event segmentation. Frame window size ($\phi$) is defined as the maximum joining window size between events; a high $\phi$ value can causes separate detected events to merge, which decreases the overall performance.

\paragraph{Frame Level}
The recall value in frame level ROC is calculated as the ratio of true positive frames (event present) to the number of positive frames in the annotations dataset, while the false positive rate is expressed as ratio of the false positive frames to the total number of negative frames (event not present) in the annotation dataset. Threshold value ($\psi$) is varied to obtain a single ROC line, while varying the frame window size ($\phi$) results in a different ROC line.

\paragraph{Activity Level}
The Hungarian matching (Munkres assignment) algorithm is utilized to achieve one to one mapping between the ground truth labeled events and the detected events. Recall is defined as ratio of the number of correctly detected events (overlapping frames) to the total number of groundtruth events. For the activity level ROC chart, the recall values are plotted against the false positive rate per minute, defined as the ratio of the total number of false positive detected events to the total duration of the dataset in minutes. The false positive rate per minute evaluation metric is also used in the ActEV TRECVID challenge \cite{ActEV}. Frame window size value ($\phi$) is varied to obtain a single ROC line, while varying the threshold value ($\psi$) results in a different ROC line.

\subsection{Ablative Studies} \label{sec:abl}
Different variations of our framework (section \ref{sec:method}) have been evaluated to quantify the effect of individual components on the overall performance. In our experiments, we tested the base model, which trains the perceptual prediction framework - including attention unit - using the prediction loss function for backpropagation of the error signal. We refer to the base model as \textit{LSTM+ATTN}. We also experimented with the effect of removing the attention unit, from the model architecture, on the overall segmentation performance; results of this variation are reported under the model name \textit{LSTM}. Further testing includes using the motion weighted loss for backpropagation of the error signal. We refer to the motion weighted model as \textit{LSTM+ATTN+MW}. Each of the models has been tested extensively; results are reported in Sections \ref{sec:quan} \& \ref{sec:qual}, as well as visually expressed in Figures \ref{fig:roc_1}, \ref{fig:roc_2}, \ref{fig:roc_3} , \ref{fig:samples1} \& \ref{fig:samples2}.

\subsection{Quantitative Evaluation} \label{sec:quan}  

We tested three different models, \textit{LSTM}, \textit{LSTM+ATTN}, and \textit{LSTM+ATTN+MW}, for frame level and activity level event segmentation. Simple and adaptive gating functions (Section \ref{sec:EG}), were applied to prediction and motion weighted loss signals (Section \ref{sec:loss}) for frame level and activity level experiments. For each model we vary parameters such as the threshold value $\psi$ and the frame window size $\phi$ to achieve the ROC charts presented in Figures \ref{fig:roc_1}, \ref{fig:roc_2} \& \ref{fig:roc_3}.

It is to be noted that thresholding a loss signal does not necessarily imply that the model was trained to minimize this particular signal. In other words, loss functions used for backpropagating the error to the models' learnable  parameters are identified only in the model name (Section \ref{sec:abl}); however, thresholding experiments have been conducted on different types of loss signals, regardless of the backpropagating loss function used for training.

The best performing model, for frame level segmentation, (\textit{LSTM+ATTN,$\psi=1000$}) is capable of achieving \{40\%, 60\%, 80\%\} frame recall value at \{5\%, 10\%, 20\%\} frame false positive rate respectively. Activity level segmentation can recall \{80\%, 90\%, 95\%\} of the activities at \{0.02, 0.1, 0.2\} activity false positive rate per minute, respectively, for the model (\textit{LSTM+ATTN,$\phi=0.0021$}) as presented in Figure \ref{fig:roc_3}. A 0.02 false positive activity rate per minute can also be interpreted as one false activity detection every 50 minutes of training (for detecting 80\% of the groundtruth activities).

Comparing the results shown in Figures \ref{fig:roc_2} \& \ref{fig:roc_3} indicate a significant increase of overall performance when using an adaptive threshold for loss signal gating. The efficacy of adaptive thresholding is evident when applied to activity level event segmentation. Results have also shown that the model can effectively generate attention maps (Section \ref{sec:qual}) without impacting the segmentation performance.

\subsection{Qualitative Evaluation} \label{sec:qual}  

\begin{figure*}
\centering
\subfigure[Samples at daytime while the bird is stationary]
{\label{fig:samples-day}
\includegraphics[width = 0.92\linewidth]{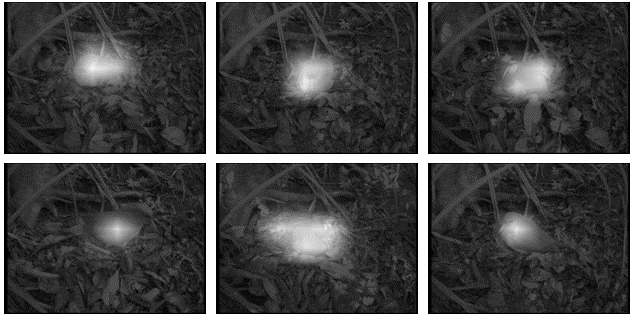}}
\subfigure[Samples at night while the bird is stationary]
{\label{fig:samples-night}
\includegraphics[width = 0.92\linewidth]{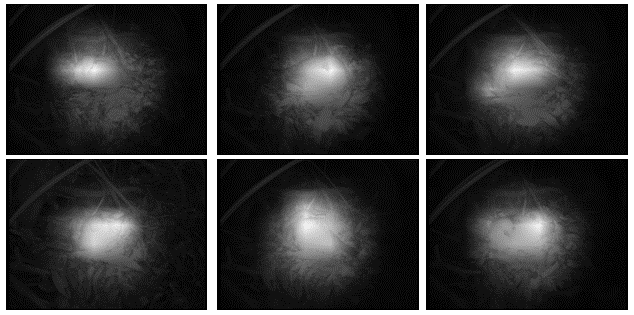}}
\caption{Samples of Bahdanau attention weights visualized on input images.}
\label{fig:samples1}
\end{figure*}

\begin{figure*}
\centering
\subfigure[Samples at daytime with moving shadows]
{\label{fig:samples-shadows}
\includegraphics[width = 0.92\linewidth]{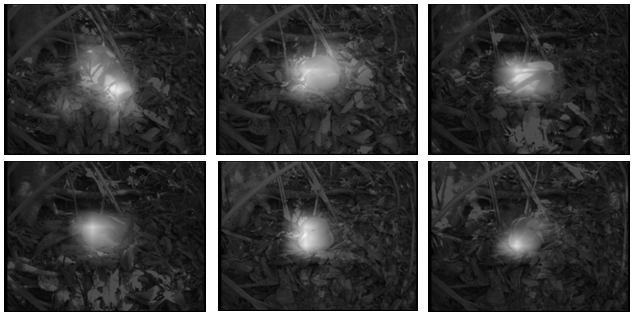}}
\subfigure[Samples at daytime while the bird is moving]
{\label{fig:samples-moving}
\includegraphics[width = 0.92\linewidth]{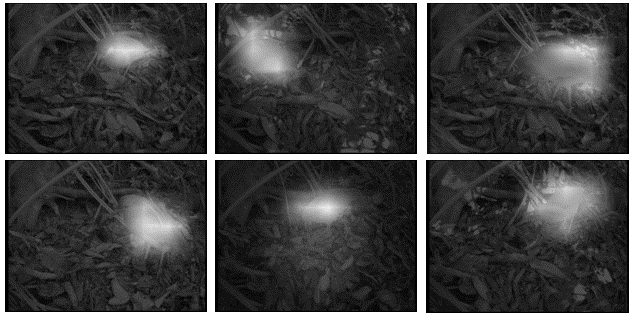}}
\caption{Samples of Bahdanau attention weights visualized on input images.}
\label{fig:samples2}
\end{figure*}

A sample of the qualitative attention results are presented in Figure \ref{fig:samples1} \& \ref{fig:samples2}. The attention mask, extracted from the model, has been trained to track the event in all processed frames. Our results show that the events are tracked and localized in various lighting (shadows, day/night) and occlusion conditions. Attention has also learned to indefinitely focus on the bird regardless of its motion state (stationary/Non-stationary), which indicates that the model has acquired a high-level temporal understanding of the events in the scene and learned the underlying structure of the bird. Supplementary results \footnote{Available at \url{https://ramyamounir.github.io/projects/EventSegmentation/}} display a timelapse of attention weighted frames during illumination changes and moving shadows. We also provide a supplementary video showing the prediction loss signal, motion weighted loss signal and attention mask during a walk in and out event (summarized in Figure \ref{fig:event}).



\section{Conclusion}
We demonstrate a self-supervised approach to temporal event segmentation. Our framework can effectively segment a long sequence of activities (video) into a set of individual events. We introduce a novel approach to extract attention results from unsupervised temporal event segmentation network. Gating the loss signal with different threshold values can result in segmentation at different granularities. Quantitative and qualitative results are presented in the form of ROC charts and attention weighted frames. Our results demonstrate the effectiveness of our approach in understanding the higher level spatiotemporal features required for practical temporal event segmentation.

\section*{Acknowledgment}
This research was supported in part by the US National Science Foundation grants CNS 1513126 and IIS 1956050. The bird video dataset used in this paper was made possible through funding from the Polish National Science Centre (grant NCN 2011/01/M/NZ8/03344 and 2018/29/B/NZ8/02312). Province Sud (New Caledonia) issued all permits - from 2002 to 2020 - required for data collection.

{\small
\bibliographystyle{IEEEtran}
\bibliography{egbib.bib}
}

\end{document}